\documentclass{article}
\usepackage{ijcai23}

\usepackage{times}
\usepackage{soul}
\usepackage{pifont} 
\usepackage{url}
\usepackage[hidelinks]{hyperref}
\usepackage[utf8]{inputenc}
\usepackage[small]{caption}
\usepackage{graphicx}
\usepackage{amsmath}
\usepackage{amsthm}
\usepackage{enumitem}
\usepackage{booktabs}
\usepackage{algorithm}
\usepackage{algorithmic}
\usepackage[switch]{lineno}
\usepackage{multirow}
\usepackage{array}
\usepackage{amsfonts}
\usepackage{cleveref}
\usepackage{subcaption}
\usepackage{xcolor}

\title{CaT-GNN: Enhancing Credit Card Fraud Detection via Causal \\ Temporal Graph Neural Networks}

\author{
Yifan Duan$^{1}$\and
Guibin Zhang$^{3}$\and
Shilong Wang$^{1}$\and
Xiaojiang Peng$^{4}$\and
Wang Ziqi$^{1}$\and
Junyuan Mao$^{1}$\and \\
Hao Wu$^{1*}$\and
Xinke Jiang$^{2*}$\and
Kun Wang$^{1}$\footnote{Corresponding Authors} \\
\affiliations{ 
$^1$University of Science and Technology of China, Hefei China, $^2$Peking University, Beijing, China, $^3$Tongji University, Shanghai, China, $^4$Shenzhen Technology University, Shenzhen, China\\
}\emails
\{Yifan Duan, Shilong Wang, Wang Ziqi, Hao Wu, wk520529, maojunyuan\}@mail.ustc.edu.cn\\
bin2003@tongji.edu.cn, pengxiaojiang@sztu.edu.cn,  thinkerjiang@foxmail.com
}

\begin{document}

\maketitle

\begin{abstract}
Credit card fraud poses a significant threat to the economy. While Graph Neural Network (GNN)-based fraud detection methods perform well, they often overlook the causal effect of a node's local structure on predictions. This paper introduces a novel method for credit card fraud detection, the \textbf{\underline{Ca}}usal \textbf{\underline{T}}emporal \textbf{\underline{G}}raph \textbf{\underline{N}}eural \textbf{N}etwork (CaT-GNN), which leverages causal invariant learning to reveal inherent correlations within transaction data. By decomposing the problem into discovery and intervention phases, CaT-GNN identifies causal nodes within the transaction graph and applies a causal mixup strategy to enhance the model's robustness and interpretability. CaT-GNN consists of two key components: Causal-Inspector and Causal-Intervener. The Causal-Inspector utilizes attention weights in the temporal attention mechanism to identify causal and environment nodes without introducing additional parameters. Subsequently, the Causal-Intervener performs a causal mixup enhancement on environment nodes based on the set of nodes. Evaluated on three datasets, including a private financial dataset and two public datasets, CaT-GNN demonstrates superior performance over existing state-of-the-art methods. Our findings highlight the potential of integrating causal reasoning with graph neural networks to improve fraud detection capabilities in financial transactions.
\end{abstract}

\section{Introduction}

The substantial damages wrought by financial fraud continue to garner ongoing focus from academic circles, the business sector, and regulatory bodies \cite{jiang2016suspicious,aleksiejuk2001simple}.  Fraudsters masquerade as ordinary users and attack transactions made with credit cards \cite{ileberi2022machine}, which may inflict substantial economic losses and pose a severe threat to sustainable economic growth \cite{alfalahi2019conceptual}. Consequently, effective detection of financial fraud is imperative for safeguarding the economy and consumer security.

In the financial deception realm, identifying credit card fraud has garnered considerable attention among both industry and academia \cite{bhattacharyya2011data}.  Traditional approaches to detecting fraudulent activities typically entail meticulous examination of each transaction for irregularities, employing predefined criteria such as verification against lists of compromised cards or adherence to established transaction thresholds \cite{maes2002credit,fu2016credit}. However, the aforementioned anti-fraud systems, based on expert prior and rules, are often susceptible to exploitation by fraudsters, who can circumvent detection by crafting ingenious transaction methods that elude the system's scrutiny of illicit activities. Toward this end, predictive modeling has been introduced, aiming to autonomously identify patterns that suggest fraudulent activity and calculate a corresponding risk score. 

\begin{figure}[tb]
    \centering
    \includegraphics[width=0.5\textwidth]{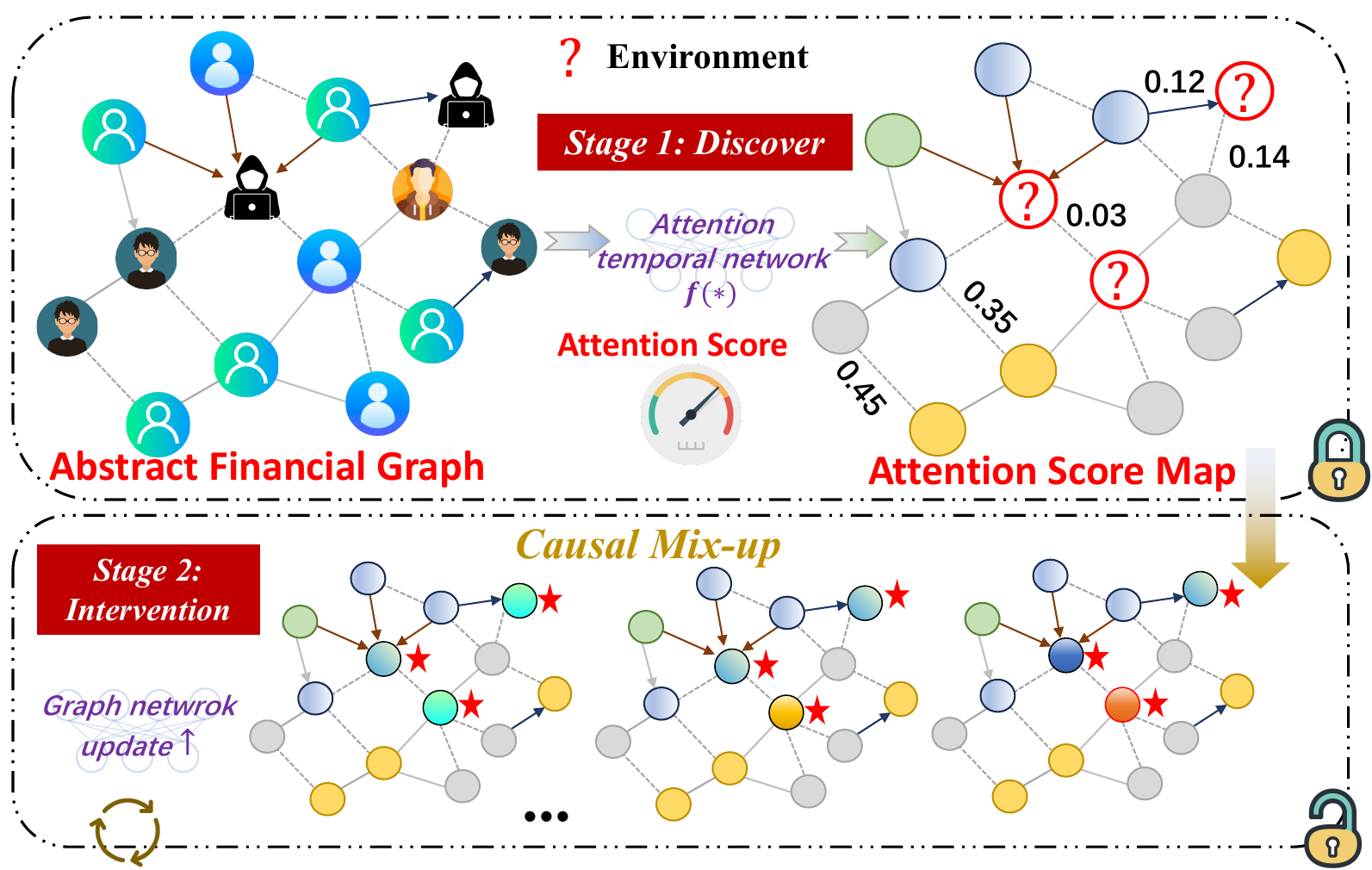}
    \vspace{-1.3em}
    \caption{The model overview. First Stage (discovery): we utilize an attention map in the attention temporal network to identify causal nodes and environment nodes. Second Stage: Intervention, we apply causal mix-up enhancement to the environment nodes.}
    \vspace{-0.2cm}
\end{figure}

Currently, state-of-the-art predictive models are focused on using deep learning methods, capturing potential illegal patterns in a data-driven manner \cite{fu2016credit,dou2020enhancing}. For instance, \cite{liu2021pick} introduces PC-GNN, a Graph Neural Network approach that effectively handles class imbalance in graph-based fraud detection by selectively sampling nodes and edges, particularly focusing on the minority class. Moreover, \cite{xiang2023semi} leverages transaction records to construct a temporal transaction graph, applying a Gated Temporal Attention Network to effectively learn transaction representations and model fraud patterns. Unfortunately, \textbf{i)} these methods often overlook the intrinsic patterns and connections within the data due to a lack of consideration for \underline{local structure consistency}; \textbf{ii)} they lack the ability to uncover the \underline{causal nature} of each specific case, which leads to inadequate differentiation between the attributes of causal nodes and environment nodes, thereby impairing the model's generalization capabilities; \textbf{iii)} they lack \underline{interpretability} in making specific predictions.

In this paper, we introduce a novel \textbf{\underline{Ca}}usal \textbf{\underline{T}}emporal \textbf{\underline{G}}raph \textbf{\underline{N}}eural \textbf{N}etwork, termed CaT-GNN, aiming at providing an interpretable paradigm for credit card fraud detection. Guided by the currently popular causal invariant learning techniques \cite{chang2020invariant,liu2022towards}, CaT-GNN's primary objective is to \textit{unveil the inherent correlations in the transaction attribute data of nodes within available temporal transaction graphs}, thereby offering interpretability for complex transaction fraud problems.

To unravel causal correlations, specifically, we decompose the algorithmic process of CAT-GNN into two stages - \textbf{discovery} and \textbf{intervention}. The goal of the discovery stage is to identify potential causal components within observed temporal graph data, where we introduce a causal temporal graph neural network for modeling. Utilizing the popular node-attention metrics \cite{velivckovic2017graph,xiang2023semi}, we employ attention score to locate key nodes, designated as causal and environment nodes. In the intervention process, we aim to reasonably enhance potential environment nodes. This approach is designed to align with and perceive the underlying distribution characteristics in explicit fraud networks, thereby boosting our temporal GNN's ability to identify and understand problematic nodes. Furthermore, drawing inspiration from \cite{wang2020nodeaug}, to ensure that causal interventions between nodes do not interfere with each other, we create parallel universes for each node. Consequently, the model is exposed to a wider potential data distribution, providing insights for fraud prediction with a causal perspective. This process can further be understood as a back-door adjustment in causal theory \cite{pearl2009causality,pearl2018book}.
The contributions of this paper are summarized as follows:

\begin{itemize}[leftmargin=*]
\item We propose a novel method, CaT-GNN, that embodies both causality and resilience for the modeling of credit card fraud detection. By harnessing causal theory, known for its interpretability, CaT-GNN enables the model to encompass a wider potential data distribution, thereby ensuring its exceptional performance in this task.

\item CaT-GNN, characterized by its refined simplicity, initially identifies causal nodes and subsequently refines the model into a causally coherent structure. It aims to achieve invariance in attribute information and temporal features through semi-supervised learning, thereby providing a bespoke and robust foundation for targeted tasks.

\item We evaluate CaT-GNN on three representative datasets, including a \textbf{\underline{private}} financial benchmark, and the other two are public settings. Extensive experiments show that our proposed method outperforms the compared state-of-the-art baselines in credit card fraud detection, thanks to the casual intervention of the node causal augment.
\end{itemize}

\section{Preliminaries}\label{preliminary}
\paragraph{\textit{Definition 1. (Multi-Relation Graph)}}
 The Multi-Relation Financial Graph $\mathcal{G}$ is defined as $\mathcal{G}=(\mathcal{V, E, X, Y})$, where $\mathcal{V}= \left\{v_{1},v_{2},\cdots, v_{N} \right\} $ represents the set of nodes, with $N=| \mathcal{V}|$ indicates the total number of nodes. $\mathcal{X} \in \mathbb{R}^{N\times d}$ denotes the node  features with $x_{i} \in \mathbb{R}^{d}$ as its entry for node $v_i$, $d$ is the feature dimension. Each node $v_{i}$ is assigned a label $y_{i} \in \mathcal{Y}$, which is a binary variable with the value in $ \left\{0,1 \right\} $. $\mathcal{E} = \left\{\mathcal{E}_{1}, \cdots, \mathcal{E}_{R} \right \}$ signifies the set of edges, partitioned into $R$ distinct types of relations.

\paragraph{\textit{Definition 2. (Graph-based Fraud Detection)}}
The graph-based fraud detection problem is defined on the multi-relation graph $ \mathcal{G} = (\mathcal{V, E, X, Y)} $. For such a problem, each node $v_i$ represents the target entity such as a transaction record, and has a label $y_{i} \in Y$, where $y_{i}=0$ represents benign and $y_{i}=1$ represents fraud. The objective of graph-based fraud detection is to identify fraud nodes that stand out distinctly from the non-fraudulent, or benign, nodes within a multi-relational graph $\mathcal{G}$. This task is effectively approached as a binary classification problem focused on nodes within the graph $\mathcal{G}$. And the surprised Binary Cross-Entropy Loss function is:$ \quad \min_{\Theta}\mathcal{L} = -\frac{1}{B}\sum_{u=1}^{B}\left(y_i^{\top} \log \left(\hat{y}_i\right)+\left(1-y_i\right)^{\top} \log \left(1-\hat{y}_i\right)\right) + \eta ||\Theta||^2,$
where $B$ is the batch size, $\hat{y}_i \in [0,1]$ is the predicted probability, and $y_i \in\{0,1\}$ is the ground truth. $\Theta$ is the parameter of the GNN predictor.

\section{Methodology}

In Section~\ref{sec: motivation}, we explore the motivation behind our approach, emphasizing the crucial role of understanding the local structure and causal relationships within transaction data to improve detection accuracy. Section~\ref{sec: Discovering and Intervention} introduces our two-phase method: discovery and intervention. Section Section~\ref{sec: Discovering and Intervention} provides the causal theory support.

\subsection{Motivation}
\label{sec: motivation}
\begin{figure}[h]
    \centering
    \includegraphics[width=0.5\textwidth]{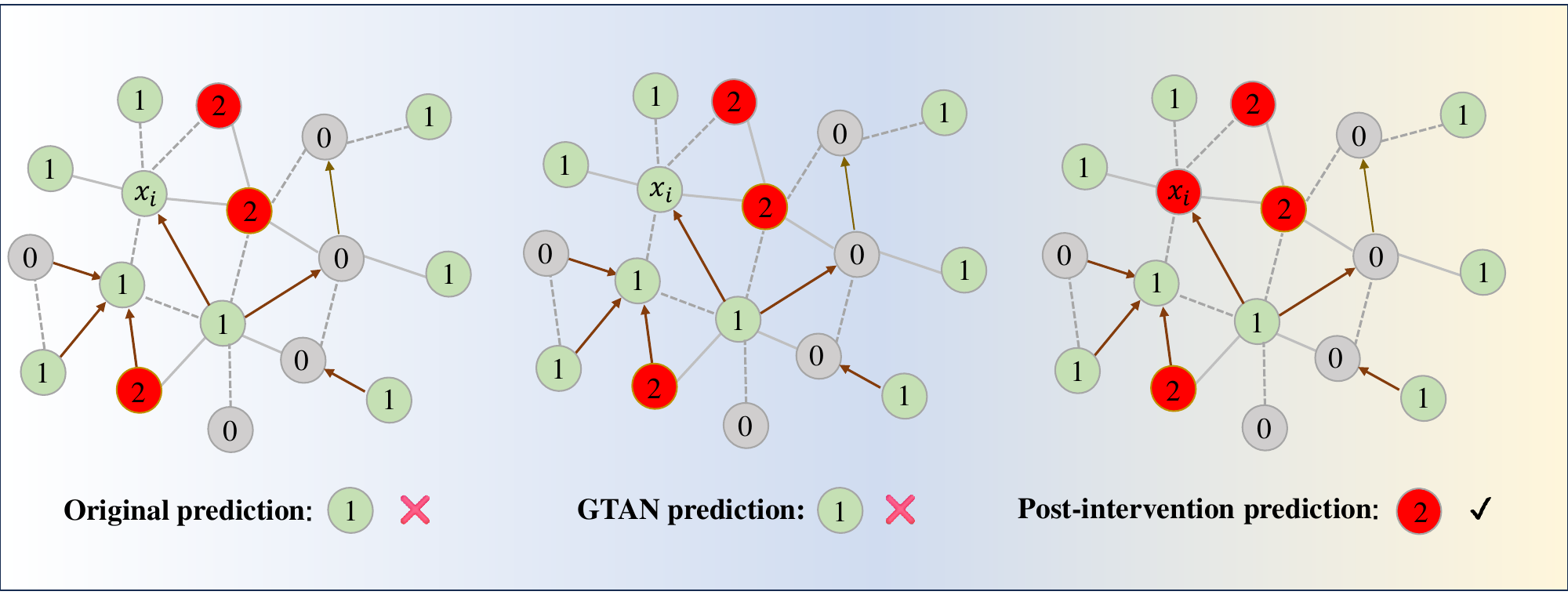}
    \vspace{-1.3em}
    \caption{\textbf{\textit{Motivation.}} The original prediction incorrectly identifies a fraudster (central node labeled \( x_i \)) as benign, as does the state-of-the-art GTAN model. Following our causal intervention, the prediction is correctly adjusted to identify \( x_i \) as a fraudster. \textit{Green: benign users, red: fraudsters, gray: unlabeled nodes.}}
    \label{fig:motivation}
    \vspace{-0.2cm}
\end{figure}

\begin{figure*}[t]
    \centering
    \includegraphics[width=\textwidth]{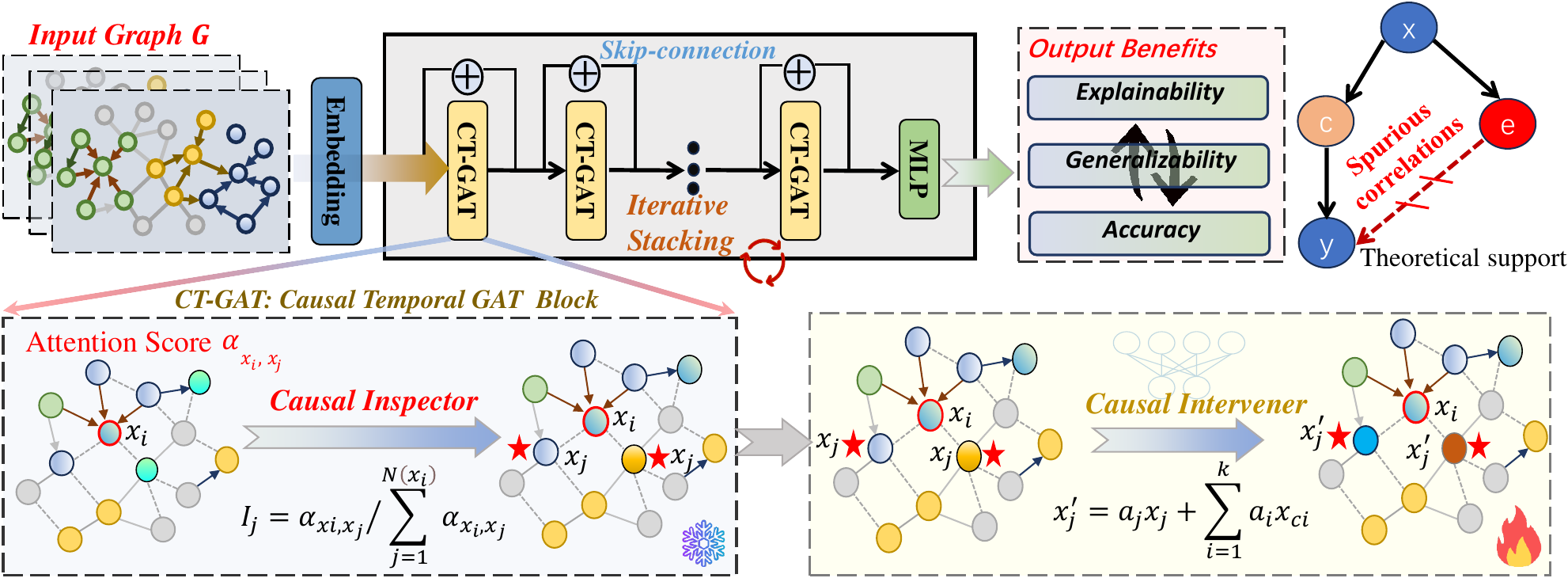}
    \vspace{-1.5em}
    \caption{The depiction of the proposed model's architecture, featuring a causal temporal graph attention mechanism, alongside the theoretical support for backdoor adjustment.}
    \label{fig:model}
    \vspace{-0.4cm}
\end{figure*}

Taking the arXiv \cite{hu2020open} dataset as an example, real-world graphs often exhibit locally variable structures, that is, the distribution of node attributes differs from the distribution of local structural properties \cite{feng2021should}. We observe that this phenomenon is also prevalent in the financial sector, where cunning fraudsters may disguise themselves through various means (such as feature camouflage and relationship disguise) to connect with users who have a good credit transaction history \cite{dou2020enhancing}. \textit{In such scenarios, if we simply aggregate node information and neighbor information together, it is likely to obscure the suspiciousness of the fraudsters, which contradicts our objective.} This situation tends to yield poorer training outcomes, especially in a semi-supervised learning environment with limited labeled data. Existing methods do not \underline{incorporate causal} factors into credit card fraud modeling, resulting in models that fail to learn the intrinsic connections of node attributes. This oversight further leads to the neglect of causal attribute structure differences on test nodes, thereby reducing the model's generalizability. By comprehensively examining the confounding variables, we are able to significantly alleviate the aforementioned issue, as illustrated in Figure \ref{fig:motivation}. This strategy is the cornerstone of our framework and is also known as the ``backdoor adjustment" technique \cite{pearl2009causality,pearl2018book}.

\subsection{Discovering \& Intervention}
\label{sec: Discovering and Intervention}
Based on the motivation, we adopt a causal perspective to analyze the attribute aggregation process and formalize principles for distinguishing between causal and non-causal elements within local structures. We first introduce the discovery process to effectively examine the causal and environment nodes within the current node's local structure. In response, we refine the temporal attention graph network mechanism\cite{xiang2022temporal} into \textit{a causal temporal GAT mechanism} as shown in the upper half of Figure \ref{fig:model}. This refinement introduces two key components designed to accurately identify both environmental and causal nodes, which enhances our ability to understand and manipulate the local structural dynamics more effectively.

 In the context of temporal transaction graphs, we maintain a set of transaction records, denoted as \(R = [r_{t1}, r_{t2}, \cdots, r_{ti}]\), alongside their embeddings \(X = [x_{t1}, x_{t2}, \cdots, x_{ti}]\) obtained via a projection layer. As demonstrated in \cite{shi2020masked}, GNNs are capable of concurrently propagating attributes and labels. Consequently, we integrate fraud labels as an embedded feature within the attribute embedding \(x_{ti}\), employing masking techniques to prevent label leakage \cite{xiang2023semi}. However, this aspect does not constitute the primary focus of our research. 
 
\textbf{Causal-Inspector:} We design a Causal-Inspector to identify causal and environment nodes as shown in the bottom left corner of Figure \ref{fig:model}. To aggregate information efficiently, we employ the aforementioned causal temporal graph attention mechanism, which allows for dynamic information flow based on the temporal relationships among transactions. Leveraging a multi-head attention mechanism, we compute temporal attention scores that serve as weights for each neighboring node, facilitating the assessment of each neighbor's causal importance, which can be formulated as follows: 
\begin{align}
\alpha_{x_i, x_j}^h = \frac{\exp \bigl(\text{LeakyReLU}(W_a^T [x_i \oplus x_j])\bigl)}{\sum_{j \in N(x_i)} \exp(\text{LeakyReLU}\bigl(W_a^T [x_i \oplus x_j])\bigl)},
\end{align}
where \( W_a \) is a learnable weight matrix, \( \alpha_{x_i, x_j} \) represents the attention weight of node \( x_i \) with respect to node \( x_j \) in one head, which determines the importance of node \( x_i \) relative to node \( x_j \). The \( \oplus \) symbol represents the concatenation operation. \( N(x_i) \) is the set of temporal neighboring nodes of node \( x_i \). In order to quantify the importance of each node \(x_j\) we aggregate the attention weights \(\alpha_{x_i, x_j}^h\) from each attention head and compute the average to determine the final weight of the node. Then, based on its final weight, we calculate its normalized importance:
\begin{align}
\bar{\alpha}_{x_i,x_j} = \frac{1}{H} \sum_{h=1}^{H} \alpha_{x_i,x_j}^h \rightarrow \hspace{0.2cm} I_j = \bar{\alpha}_{x_i,x_j}/\sum_{j=1}^{N(x_i)} \bar{\alpha}_{x_i,x_j}
\end{align}
where $H$ is the total number of attention heads and $I_i=[I_{j1},\cdots, I_{jN(x_i)}]$ represents the set of importance of each node with respect to $x_i$. This formula calculates the normalized importance weight $I_j$, representing the importance of node $x_j$ by compiling the contributions from all attention heads, thus providing a comprehensive measure of node significance.
To segregate the nodes into environmental and causal categories, we introduce a proportion parameter \( r_e \), ranging between 0 and 1, which denotes the fraction of nodes to be earmarked as environment nodes. This approach affords us the flexibility to select environment nodes tailored to the specific exigencies of the graph. We use the $\text{argmin}(\cdot)$ function to select the \( \lceil r_eN \rceil \) nodes with the lowest importance scores as environment nodes. Therefore, a ranking function \( R \) is defined to map \( I_j \) to its rank among all node importance scores. Then, we determine the environment set \( S_e \) as:
\begin{align}
{S_e} = \bigl\{x_j |\mathop{\text{argmin} }\limits_{j=1,\cdots,\lfloor r_eN \rfloor} R(I_j) \bigl\}.
\end{align}

\noindent The remaining nodes, those not in \( {S_e} \), naturally form the set of causal nodes \( S_c \).This method ensures that nodes with the lowest importance scores are precisely selected as environmenta nodes according to the proportion \( r_e \), while the rest serve as causal nodes. Due to the differences between test and training distributions \cite{feng2021should}, CaT-GNN is dedicated to perceiving the essence of temporal graph data, thereby enhancing generalization capabilities and robustness.

\textbf{Causal-Intervener:} We design a Causal Intervener as shown in the bottom right corner of Figure \ref{fig:model}, which employs a transformative mixup strategy known as a causal mixup, that blends environment nodes with a series of causally significant nodes. Given an environmental node \( x_j \in S_e\), We select the causal nodes \( \{x_{c1}, x_{c2}, \cdots, x_{ck}\} \) with the highest importance scores, which are computed as outlined in the Causal-Inspector, from the causal set $S_c$ at a proportion of $r_c$. The \emph{causal mixup} is then executed by linearly combining the environmental node with the selected causal nodes, weighted by their respective coefficients
\( \{a_j, a_1, a_2, \cdots, a_k\} \), which are learned through a dedicated linear layer:
\begin{align}
x'_j = a_j x_j + \sum_{i=1}^{k} a_i x_{ci},
\end{align}
\vspace{-0.2em}
where \( x'_j \) is the causally mixed environmental node, $k$ is the number of selected causal nodes, \( a_j \) is the self-weight of the environmental node reflecting its inherent causal significance, and \( a_i \) is the causal node weight. These weights are normalized such that \( a_j + \sum_{i=1}^{k} a_i = 1 \). The incorporation of the causal mixup enhances the robustness of the model against distributional shifts by embedding a richer causal structure within the environmental node. By adapting the causal structure to the environmental context, the Causal-Intervener aims to mitigate the disparity between training and test distributions, thus bolstering the model's generalizability. Finally, we aggregate the information, and the outputs of multiple attention heads are concatenated to form a more comprehensive representation:
\vspace{-0.5em}
\begin{equation}\footnotesize
\begin{aligned}
& \mathcal{H} = \sum_{x_i \in \mathcal{X}} \sigma \bigl( \sum_{x_j \in N(x_i)} \alpha_{x_i, x_j} x_j'\bigl), \\
& \mathcal{M} = \bigl(\mathcal{H}_1 \oplus \cdots \oplus  \mathcal{H}_\text{h}\bigl) W_c,
\end{aligned}
\end{equation}
\vspace{-0.2em}
where $W_c$ is a learnable weight matrix, $\mathcal{H}$ is a attention head, $\mathcal{M}$ denotes the aggregated embeddings. It is important to highlight that the causal intervention result $x_j'$ on an environmental node $x_j$ with respect to $x_i$ is essentially a duplicate of $x_j$ and does not modify $x_j$ itself. This distinction is crucial as it guarantees that the process of augmenting central nodes within individual local structures remains mutually non-disruptive. By preserving the original state of $x_j$, we ensure that enhancements applied to central nodes in one local structure do not adversely affect or interfere with those in another, maintaining the integrity and independence of local structural enhancements~\cite{wang2020nodeaug}.

\subsection{Causal Support of CaT-GNN}
\label{sec: Causal Support of CaT-GNN}
In elucidating the causal backbone of CaT-GNN, we invoke causal theory to formulate a Structural Causal Model (SCM) as propounded by \cite{pearl2009causality}. This framework scrutinizes four distinct elements: the inputs node attribute $ X $, the truth label $ Y $ decided by both the attribute of causal nodes of \( X \) symbolized as \( \tilde{C} \), and the confounder \( E \), emblematic of the attribute of environment nodes. The causal interplay among these variables can be articulated as follows:

\begin{itemize}[leftmargin=*]
  \item $\tilde{C} \xleftarrow{} $X$ \xrightarrow{} \ $E$ $. The local structure of node attribute \( X \) is composed of causal nodes attributes \( \tilde{C} \) and environment nodes attributes $E$. 
  \item $ \tilde{C} \xrightarrow{} \ $Y$ \ \xleftarrow{} $E$ $. The causal attributes  \( \tilde{C} \) actually determine the true value $Y$, however, the environmental attributes $E$ also affect the prediction results, causing spurious associations.

\end{itemize}

\noindent Do-calculus \cite{pearl2009causality} is a trio of rules within the causal inference framework that facilitates the mathematical deduction of causal effects from observed data. These rules enable manipulation of $\text{do}(\cdot)$ operator expressions, essential for implementing interventions in causal models:
\begin{equation}
\begin{aligned}
& P\bigl(Y|\text{do}(\hat{C}), E\bigl) = P(Y|\hat{C}),
 \\
& P\bigl(Y|\text{do}(\hat{C}), \text{do}(E)\bigl) = P\bigl(Y|\text{do}(\hat{C}), E\bigl),
\\
& P\bigl(\hat{C}|\text{do}(Y)\bigl) = P\bigl(\hat{C}|Y\bigl).
\end{aligned}
\end{equation}

\noindent Typically, a model \( M_{\theta} \) that is trained using Empirical Risk Minimization (ERM) may not perform adequately when generalizing to test data distribution \( P_{test} \). These shifts in distribution are often a result of changes within environment nodes, necessitating the need to tackle the confounding effects. As illustrated in Figure \ref{fig:model}, we apply causal intervention to enhance the model's generalizability and robustness. To this end, our approach utilizes do-calculus \cite{pearl2009causality} on the variable \( C \) to negate the influence of the backdoor path \( E \rightarrow Y \) by estimating \( P\bigl(Y|\text{do}(\hat{C})\bigl) = P_m(Y|\hat{C}) \):
\vspace{-0.5em}
\begin{equation}
\footnotesize
\begin{aligned}
P\bigl(Y|\text{do}(\hat{C})\bigl) &= \sum_{i}^{N_e} P\bigl(Y|\text{do}(\hat{C}), E=E_i\bigl) P\bigl(E=E_i|\text{do}(\hat{C})\bigl) \\ 
&= \sum_{i}^{N_e} P\bigl(Y|\text{do}(\hat{C}), E=E_i\bigl) P(E=E_i) \notag \\
&= \sum_{i}^{N_e} P\bigl(Y|\hat{C}, E=E_i) P(E=E_i),
\end{aligned}
\end{equation}
where \( N_e \) signifies the count of environment nodes, with \( E_i \) indicating the \( i \)-th environmental variable. The environmental enhancement of Cat-GNN is in alignment with the theory of backdoor adjustment, thereby allowing for an effective exploration of possible test environment distributions.

\section{Experiments}\label{stylefiles}
In this section, we critically assess the CaT-GNN model on a series of research questions (RQs) to establish its efficacy in graph-based fraud detection tasks. The research questions are formulated as follows:
\begin{itemize}[leftmargin=*]
  \item \textbf{RQ1}: Does CaT-GNN outperform the current state-of-the-art models for graph-based anomaly detection?
  \item \textbf{RQ2:} What is the effectiveness of causal intervention in the aggregation of neighboring information?
  \item \textbf{RQ3:} What is the performance with respect to different environmental ratios $r_e$?
  \item \textbf{RQ4:} Is CaT-GNN equally effective in semi-supervised settings, and how does it perform with limited labeled data?
  \item \textbf{RQ5:} Does the causal intervention component lead to a significant decrease in efficiency?

\end{itemize}

\subsection{Experimental Setup}

\paragraph{Datasets.} we adopt one open-source $\mathbf{f}$inacial $\mathbf{f}$raud $\mathbf{s}$emi-supervised $\mathbf{d}$ataset \cite{xiang2023semi}, termed S-FFSD\footnote{https://github.com/AI4Risk/antifraud}, with the partially labeled transaction records. Same with the definition in section \ref{preliminary}, if a transaction is reported by a cardholder or identified by financial experts as fraudulent, the label $y_{v}$ will be 1; otherwise, $y_{v}$ will be 0. In addition, we also validate on two other public fraud detection datasets $\mathbf{Yelpchi}$ and $\mathbf{Amazon}$. $\mathbf{Yelpchi}$ \cite{rayana2015collective} compiles a collection of hotel and restaurant reviews from Yelp, in which nodes represent reviews. And there are three kinds of relationship edges among these reviews. $\mathbf{Amazon}$: The Amazon graph \cite{mcauley2013amateurs} comprises reviews of products in the musical instruments category, in which nodes represent users, and the edges are the corresponding three kinds of relationships among reviews. The statistics of the above three datasets are shown in \Cref{tab:datasets}.

\begin{table}
\footnotesize
	\caption{Statistics of the three datasets.}
        \vspace{-0.3cm}
	\label{tab:datasets}
	\centering
	\begin{tabular}{ccccc}
		\toprule
		Dataset&\#Node&\#Edge&\#Fraud&\#benigh \\
		\midrule
		YelpChi&45,954&7,739,912&6,677&39,277\\[2pt]
		Amazon&11,948&8,808,728&821&11,127\\[2pt]
	    S-FFSD&130,840&3,492,226&2,950&17,553\\
		\bottomrule
	\end{tabular}
\end{table}

\paragraph{Baselines.} To verify the effectiveness of our proposed CaT-GNN, we compare it with the following state-of-the-art methods. \ding{182} \textit{Player2Vec}. Attributed Heterogeneous Information Network Embedding Framework \cite{zhang2019key}. \ding{183} \textit{Semi-GNN}. A semi-supervised graph attentive network
for financial fraud detection that adopts the attention mechanism to aggregate node embed
dings across graphs \cite{wang2019semi}. \ding{184} \textit{GraphConsis}. The GNN-based fraud detectors aim at the inconsistency problem \cite{liu2020alleviating}. \ding{185} \textit{GraphSAGE}. The inductive graph learning model is based on a fixed sample number of the neighbor nodes \cite{hamilton2017inductive}. \ding{186} \textit{CARE-GNN} The camouflage-resistant GNN-based model tackling fraud detection \cite{dou2020enhancing}. \ding{187} \textit{PC-GNN}. A GNN-based model to address the issue of class imbalance in graph-based fraud detection \cite{liu2021pick}. \ding{188} \textit{GTAN}. A semi-supervised GNN-based model that utilizes a gated temporal attention mechanism to analyze credit card transaction data \cite{xiang2023semi}. \ding{189}: \textit{CaT-GNN (PL)}. This variant of the CaT-GNN framework selects environment nodes based on a proportion $r_e$ and determines mixup weights $a_i$ via a learnable linear layer. \ding{190}: \textit{CaT-GNN (PI)}. This version employs a proportional selection of environment nodes and leverages the nodes' importance scores to inform mixup weights $a_i = I_i /\sum_{i=1}^{k} I_i$. \ding{191}: \textit{CaT-GNN (FL)}. This variant uses a fixed number of environment nodes. Mixup weights are determined by a learnable linear layer. \ding{191}:\textit{CaT-GNN (FI)}. Combining fixed environmental node selection with importance-based weighting for mixup.

\paragraph{Reproducibility}

In our experiment, the learning rate $l_{r}$ is set to 0.003, and the batch size batch $N_{batch}$ is established at 256. Moreover, the input dropout ratio $r_{dropout}$ is determined to be 0.2, with the number of attention heads $N_{head}$ set to 4, and the hidden dimension $d$ to 256. We employed the Adam optimizer to train the model over $N_{epoch}=100$ epochs, incorporating an early stopping mechanism to prevent overfitting. In GraphConsis, CARE-GNN, PC-GNN and GTAN, we used the default parameters suggested by the original paper. In Semi-GNN and Player2Vec, We set the learning rate to 0.01. In YelpChi and Amazon datasets, the train, validation, and test ratio are set to be 40\%, 20\%, and 40\% respectively. In the S-FFSD dataset, we use the first 7 months' transactions as training data, and the rest as test data. Similar to previous work~\cite{liu2021pick}, we repeat experiments with different random seeds 5 times and we report the average and standard error.
Experimental results are statistically significant
with $p < 0.05$.
Cat-GNN and other baselines are all implemented in Pytorch 1.9.0 with Python 3.8. All the experiments are conducted on Ubuntu 18.04.5 LTS server with 1 NVIDIA Tesla V100 GPU, 440 GB RAM. 

\paragraph{Metrics.} We selected three representative and extensively utilized metrics: \textbf{AUC} (Area Under the ROC Curve), \textbf{F1-macro} and \textbf{AP} (averaged precision). The first metric AUC is the area under the ROC Curve and as a single numerical value, AUC succinctly summarizes the classifier's overall performance across all thresholds. The second metric F1-macro is the macro average of F1 score which can be formulated as $F1_{macro} = 1/(l \sum_{i=1}^{l} \frac{2 \times P_i \times R_i}{P_i + R_i})$, and the third metric AP is averaged precision that can be formulated as $ AP = \sum_{i=1}^{l} (R_i - R_{i-1})P_i $, where $P_i$ stands for the Precision and $R_i$ stands for recall.

\begin{table*}[t]
\setlength{\tabcolsep}{1.1mm}
    \centering
    \small
    \caption{Performance Comparison (in percent ± standard deviation) on YelpChi, Amazon and S-FFSD datasets across five runs. The best performances are marked with \textbf{bold font}, and the second-to-best are shown \underline{underlined.}}
    \vspace{-0.3cm}
    \label{tab:performance}
    \setlength\tabcolsep{3pt}
     \resizebox{1\linewidth}{!}{
    \begin{tabular}{lccccccccc}
    \toprule
    Dataset & \multicolumn{3}{c}{YelpChi} & \multicolumn{3}{c}{Amazon} & \multicolumn{3}{c}{S-FFSD} \\
    \cline{2-4} \cline{5-7} \cline{8-10}
     Metric & AUC & F1 & AP & AUC & F1 & AP & AUC & F1 & AP \\
    \midrule
    Player2Vec & 0.7012$\pm$0.0089 & 0.4120$\pm$0.0142 & 0.2477$\pm$0.0161 & 0.6187$\pm$0.0152 & 0.2455$\pm$0.0091 & 0.1301$\pm$0.0117 & 0.5284$\pm$0.0101 & 0.2149$\pm$0.0136 & 0.2067$\pm$0.0155 \\
    Semi-GNN & 0.5160$\pm$0.0154 & 0.1023$\pm$0.0216 & 0.1809$\pm$0.0205 & 0.7059$\pm$0.0211 & 0.5486$\pm$0.0105 & 0.2248$\pm$0.0142 & 0.5460$\pm$0.0125 & 0.4393$\pm$0.0152 & 0.2732$\pm$0.0207 \\
    GraphSAGE & 0.5414$\pm$0.0029 & 0.4516$\pm$0.0954 & 0.1806$\pm$0.0866 & 0.7590$\pm$0.0053 & 0.5926$\pm$0.0087 & 0.6597$\pm$0.0079 & 0.6534$\pm$0.0095 & 0.5396$\pm$0.0101 & 0.3881$\pm$0.0089 \\
    GraphConsis & 0.7046$\pm$0.0287 & 0.6023$\pm$0.0195 & 0.3269$\pm$0.0186 & 0.8761$\pm$0.0317 & 0.7725$\pm$0.0319 & 0.7296$\pm$0.0301 & 0.6554$\pm$0.0412 & 0.5436$\pm$0.0376 & 0.3816$\pm$0.0341 \\
    CARE-GNN & 0.7745$\pm$0.0281 & 0.6252$\pm$0.0091 & 0.4238$\pm$0.0151 & 0.8998$\pm$0.0925 & 0.8468$\pm$0.0085 & 0.8117$\pm$0.0114 & 0.6589$\pm$0.1078 & 0.5725$\pm$0.0096 & 0.4004$\pm$0.0090 \\
    PC-GNN & 0.7997$\pm$0.0021 & 0.6429$\pm$0.0205 & 0.4782$\pm$0.0194 & 0.9472$\pm$0.0019 & 0.8798$\pm$0.0084 & 0.8442$\pm$0.0096 & 0.6707$\pm$0.0031 & 0.6051$\pm$0.0230 & 0.4479$\pm$0.0210\\
    GTAN & 0.8675$\pm$0.0036 & 0.7254$\pm$0.0197 & 0.6425$\pm$0.0154 & 0.9580$\pm$0.0014 & 0.8954$\pm$0.0095 & 0.8718$\pm$0.0083 & 0.7496$\pm$0.0041 & 0.6714$\pm$0.0089 & 0.5709$\pm$0.0097\\
    \midrule
    Cat-GNN(FI) & 0.8721$\pm$0.0044 & 0.7336$\pm$0.0295 & 0.6528$\pm$0.0209 & 0.9643$\pm$0.0026 & 0.9011$\pm$0.0129 & 0.8794$\pm$0.0102 & 0.7643$\pm$0.0078 & 0.6907$\pm$0.0198 & 0.5925$\pm$0.0174\\
    Cat-GNN(FL) & \underline{0.8910$\pm$0.0026} & 0.7692$\pm$0.0182 & 0.6687$\pm$0.0135 & \underline{0.9705$\pm$0.0016} & \underline{0.9125$\pm$0.0099} & \underline{0.8942$\pm$0.0081} & 0.8023$\pm$0.0067 & 0.7031$\pm$0.0154 & 0.6145$\pm$0.0169\\
    Cat-GNN(PI) & 0.8895$\pm$0.0041 & \underline{0.7706$\pm$0.0223} & \underline{0.6701$\pm$0.0181} & 0.9669$\pm$0.0021 & 0.9077$\pm$0.0113 & 0.8896$\pm$0.0095 & \underline{0.8145$\pm$0.0061} & \underline{0.7096$\pm$0.0149} & \underline{0.6294$\pm$0.0166}\\
    Cat-GNN(PL) & \textbf{0.9035$\pm$0.0035} & \textbf{0.7783$\pm$0.0209} & \textbf{0.6863$\pm$0.0127} & \textbf{0.9706$\pm$0.0015} & \textbf{0.9163$\pm$0.0104} & \textbf{0.8975$\pm$0.0089} & \textbf{0.8281$\pm$0.0054} & \textbf{0.7211$\pm$0.0115} & \textbf{0.6457$\pm$0.0156} \\
    \bottomrule
    \end{tabular}}
\end{table*}

\subsection{Performance Comparison (RQ1)}
In the experiment of credit card fraud detection across three distinct datasets, Cat-GNN showcases superior performance metrics compared to its counterparts. \underline{First of all}, Cat-GNN achieves the highest AUC in all three datasets, with values of 0.9035, 0.9706, and 0.8281 for YelpChi, Amazon, and S-FFSD, respectively. This indicates that \textbf{Cat-GNN consistently outperforms other methods in distinguishing between classes across diverse datasets}. \underline{Focusing on the F1 Score}, which balances the precision $P_i$ and recall $R_i$, Cat-GNN again tops the charts with scores of 0.7783, 0.9163, and 0.7211 for YelpChi, Amazon, and S-FFSD. This reflects the model's robustness in achieving high precision while not compromising on recall, which is essential where both false positives and false negatives carry significant consequences. \underline{Finally}, Cat-GNN's superiority extends to the AP metric, with the improvement of at least 6.82\%, 2.86\%, and 13.10\% for YelpChi, Amazon and S-FFSD respectively.

The comparative performance of Cat-GNN is particularly significant when contrasted with previous methods such as Player2Vec, Semi-GNN, and GraphSAGE. For the Amazon dataset, existing state-of-the-art models, like CARE-GNN, PC-GNN, and GTAN, have already proven effective at capturing the inherent correlations within the data. In this context, the benefits of causal intervention may not be as pronounced, possibly due to the dataset’s simpler local structures and more uniform distribution. However, for the S-FFSD dataset, our methodology exhibits significant performance improvements. This enhancement is attributed to the complex local structures and the prevalence of unlabeled nodes within the dataset. In such scenarios, causal intervention adeptly learns the inherent attribute connections, thereby boosting the model's generalization. Additionally, learning mixup weights with a linear layer is more reasonable than weighting with importance scores. Similarly, selecting environment nodes based on proportions is more sensible than choosing a fixed number of environment nodes, and the effect is also slightly better. All in all, This superior performance can be \textit{ascribed to the integration of causal theory within the Cat-GNN}, enhancing its capacity to comprehend the inherent principles of graph attributes, allowing it to discern complex patterns and interactions that other models are unable to effectively capture.

\subsection{Ablation Study  (RQ2)}

In this section, we evaluate the effectiveness of causal interventions in the aggregation within graph structures. Initially, we explore a variant without any causal intervention, termed N-CaT, which aggregates all neighboring information indiscriminately. Secondly, we introduce D-CaT, a method that omits environment nodes entirely during the aggregation phase, and directly aggregates all neighboring information in the learning process. Finally, our proposed method, CaT, integrates a causal intervention approach, simultaneously considering both causal nodes and environment nodes during aggregation to refine the learning representations.

\begin{figure*}[t]
    \centering
\includegraphics[width=0.9\textwidth]{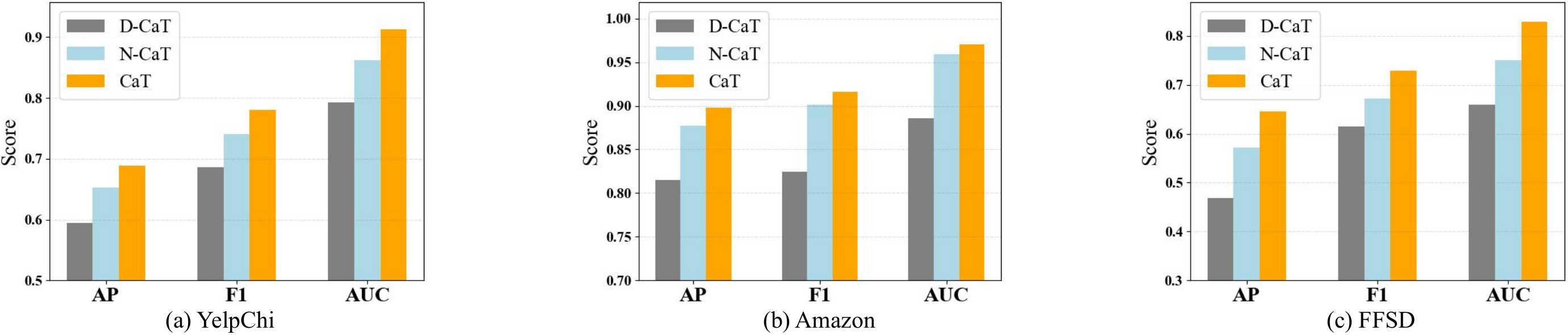}
    \vspace{-0.2cm}
    \caption{The ablation study results on three datasets. Gray bars represent the D-CaT variant, blue bars represent the N-CaT variant, and orange bars represent the CaT-GNN model.}
    \label{fig:ablation}
    \vspace{-0.3cm}
\end{figure*}

The results shown in Figure \ref{fig:ablation} highlight the importance of causal intervention in information aggregation. N-CaT, which lacks causal discernment, performs worse across all datasets compared to CaT because it does not account for causal relationships. D-CaT, which simply deletes environmental factors, shows a significant drop in performance, as the mere deletion of environment nodes prevents the model from fully learning valuable information. Our CaT method consistently outperforms the other variants across all datasets, achieving the highest AUC scores. This superior performance underscores the value of our causal intervention technique, which effectively balances the influence of causal and environment nodes, resulting in a more generalizable model.

\subsection{Parameter Sensitivity Analysis  (RQ3, RQ4)}

\begin{figure}[htb]
  \centering
\hspace{-0.2cm}
\includegraphics[scale=0.249]{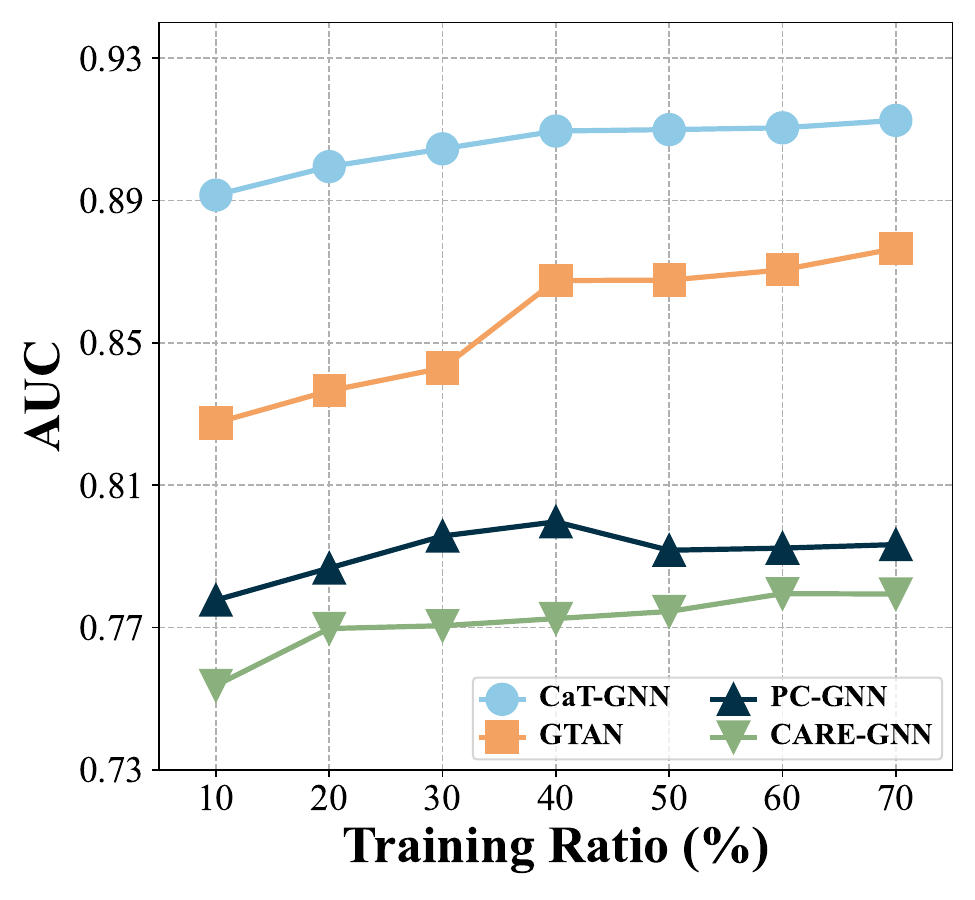}
  \includegraphics[scale=0.249]{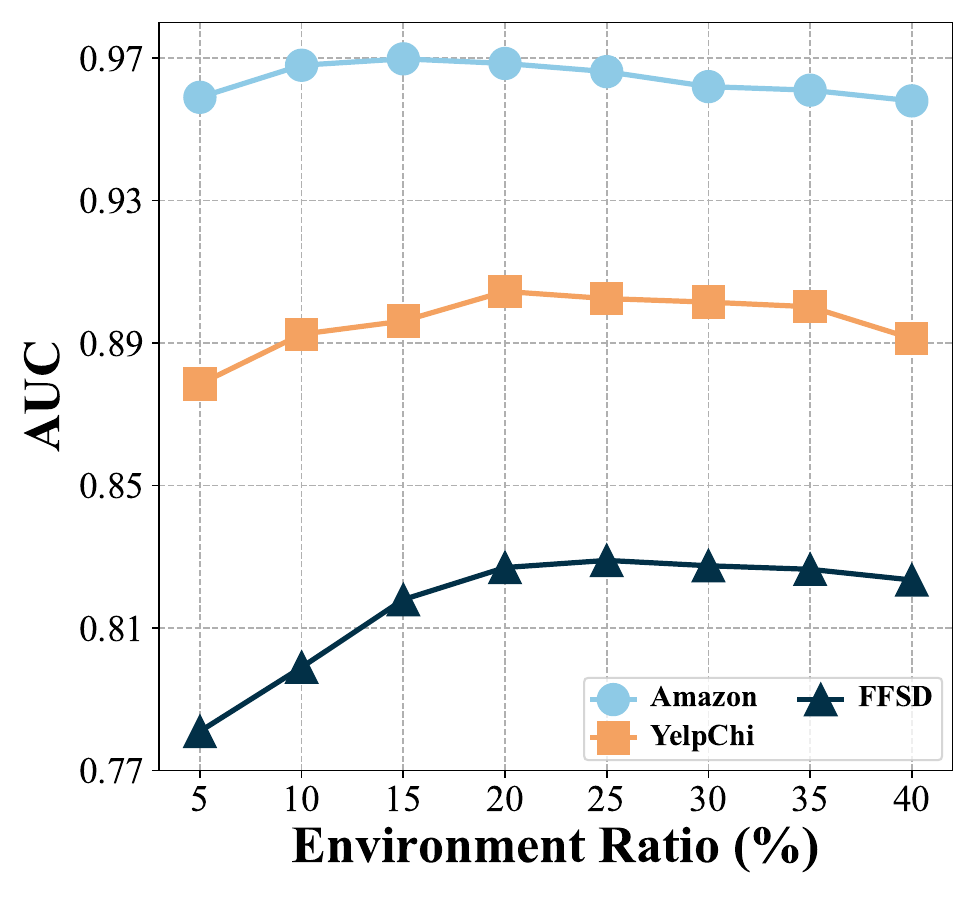}
  \vspace{-1.0em}
  \caption{Sensitivity analysis with respect to different training ratios (\textbf{Left}) and environment ratios (\textbf{Right}).
  }
   \label{fig:test}
\end{figure}

In this section, we study the model parameter sensitivity with respect to the environment nodes ratio and the training ratio. The corresponding results are reported in Figure \ref{fig:test}.

As demonstrated in the left of Figure \ref{fig:test}, using the YelpChi dataset as an example, the performance of Cat-GNN (measured by AUC as the performance metric) significantly surpasses other competitive models, including PC-GNN and CARE-GNN, across all training ratios, from 10\% to 70\%. Particularly at lower training ratios (such as 10\%), Cat-GNN remains effective for semi-supervised learning and exhibits more robust performance compared to other models.

In our sensitivity analysis of the environmental ratio as demonstrated in the right of Figure \ref{fig:test}, we observed that Cat-GNN's performance on the Amazon dataset is less affected by variations in the training ratio, with AUC fluctuations not exceeding 2\%. Conversely, on the S-FFSD dataset, as the training ratio increases from 5\% to 40\%, there is a larger fluctuation in Cat-GNN's performance. This can be attributed to the characteristics of the dataset or the differences in the distribution of labeled data.

\subsection{Model Efficiency  (RQ5)}

In this section, we present a comprehensive analysis of the efficiency of CaT-GNN. Our causal intervention aims to boost performance while maintaining computational efficiency. Table \ref{tab:effectiveness} shows that the performance enhancements are achieved without imposing significant additional computational costs. The results indicate that the execution time with causal intervention experienced only a marginal increase. This negligible rise in time is a testament to the algorithm's ability to retain its computational efficiency while incorporating our advancements. Thus, our algorithm stands as a robust solution that can cater to the needs of high-performance computing while facilitating enhancements that do not compromise on efficiency.

\begin{table}[ht]
	\caption{Experimental run times with and without causal intervention on three datasets. The experiments were conducted on a Tesla V100 40GB GPU, with the execution times measured in seconds.}
 \vspace{-1.0em}
\label{tab:effectiveness}
	\centering
	\resizebox{\linewidth}{!}{%
	\begin{tabular}{llll}
		\toprule
		Dataset & YelpChi & Amazon & S-FFSD  \\ \midrule
		No-intervention & 126.676 & 110.518 & 208.085 \\[2pt]
		Causal-intervention & 129.481 \textcolor{blue}{(+2.21\%)} & 113.660 \textcolor{blue}{(+2.84\%)} & 213.341 \textcolor{blue}{(+2.52\%)} \\[2pt]
		\bottomrule
	\end{tabular}%
	}
     \vspace{-1.0em}
\end{table}

\vspace{-0.3em}
\section{Related Works}
\vspace{-0.3em}

\paragraph{Graph Neural Network and its Variants.} Graph neural networks have been widely used in structured data prediction \cite{abadal2021computing,wu2020comprehensive,katz}~by integrating graph structure and attribute. With the development of GNNs, there are several types of GNNs nowadays. \textbf{1)}: \textit{\textbf{Recurrent Graph Neural Networks}} (RecGNNs) aim to learn node representations with recurrent neural architectures: 
\cite{scarselli2008graph,gallicchio2010graph,li2015gated,dai2018learning}. \textbf{2)}: \textit{\textbf{Convolutional Graph Neural Networks}} (ConvGNNs) generalize the operation of convolution from grid data to graph data: \cite{li2018adaptive,zhuang2018dual,xu2018powerful,chiang2019cluster}. \textbf{3)}: \textit{\textbf{Spatial–Temporal Graph Neural Networks}} (STGNNs) aim to learn complex hidden patterns from spatial-temporal graphs: \cite{yan2018spatial,wu2019graph,guo2019attention,10.1145/3583780.3615215,SPGCL}.

\paragraph{Machine-learning based credit card fraud detection.} Recently, numerous efforts have been dedicated to integrating machine learning methodologies into the research of credit card fraud detection. For instance, \cite{maes2002credit} successfully applied Bayesian Belief Networks and MLP to the Europay International dataset. \cite{csahin2011detecting} utilized decision trees and support vector machines on data from a major national bank. \cite{fu2016credit} demonstrates that convolutional neural networks outperform traditional approaches in pattern recognition for higher accuracy. However, their models were limited as they only considered individual transactions or cardholders, missing out on the potential of unlabeled data in real-world transactions \cite{xiang2023semi}.

\paragraph{GNN-based credit card fraud detection.} More recently, the focus has shifted towards graph-based approaches as the use of graph convolutional networks on datasets with partial labels has been effective for predicting node attributes within citation networks so that many GNN-based fraud detectors have been proposed to detect fraud \cite{wang2019semi,liu2018heterogeneous}. Concretely, \cite{dou2020enhancing} introduces CARE-GNN for fraud detection on relational graphs, while \cite{liu2021pick} develops PC-GNN for managing imbalanced learning on graphs. Additionally, \cite{fiore2019using} presents a generative adversarial network to enhance classification capabilities. \cite{cheng2020graph} suggested a joint feature learning model, concentrating on spatial and temporal patterns. However, these methods often lack the capability to uncover the \textbf{causal nature} of each specific case and are easily influenced by the surrounding neighbors due to the aggregation mechanism inherent in GNNs. Towards this end, we \textit{take the first step} to propose a structural causal GNN model, which introduces causal intervention and data augmentation mechanism into the aggregation process of GNN.

\section{Conclusion \& Future Work}
In this work, we introduce the Causal Temporal Graph Neural Network (CaT-GNN), a causal approach in the domain of credit card fraud detection. Our model innovates by integrating causal learning principles to discern and leverage the intricate relationships within transaction data. We validate the effectiveness of CaT-GNN through comprehensive experiments on diverse datasets, where it consistently outperforms existing techniques. Notably, CaT-GNN not only enhances detection accuracy but also maintains computational efficiency, making it viable for large-scale deployment. Future directions will explore extending this methodology to a broader range of fraudulent activities, with the aim of fortifying the integrity of financial systems globally.

\bibliographystyle{named}

\bibliography{main}
\appendix
\end{document}